\documentclass[letterpaper, 10pt, conference]{ieeeconf} 

\IEEEoverridecommandlockouts                             
\usepackage{graphics, graphicx}
\usepackage[USenglish]{babel}
\usepackage{lineno}
\usepackage{xcolor}
\usepackage{microtype}
\usepackage{eucal}
\usepackage{textcomp}
\usepackage{environ}
\usepackage{amsmath, amssymb, mathrsfs, stmaryrd} 
\usepackage{algorithm}
\usepackage{tabularx}
\usepackage{balance}
\usepackage{booktabs, multirow}
\usepackage[separate-uncertainty, digitsep]{siunitx}
\usepackage{lipsum}

\usepackage{tikz}
\usepackage{tabularx}
\usepackage{pgfplots,pgfplotstable}
\usepgfplotslibrary{fillbetween}
\usepgfplotslibrary{external}
\pgfplotsset{compat = 1.5}

\definecolor{ais-blue}{rgb}{0.2, 0.4, 0.6}
\definecolor{ais-red}{rgb}{0.6, 0., 0.}
\definecolor{ais-black}{rgb}{0., 0., 0.}
\definecolor{ais-gray}{rgb}{0.5, 0.5, 0.5}

\newcommand{\etal}{\emph{et~al. }}

\newcommand{\Bel}{bel}
\newcommand{\Measurements}{\mathbf{z}}
\newcommand{\Measurement}{z}
\newcommand{\Pose}[1]{\mathbf{#1}}

\newcommand{\px}{\textrm{px}}

\newcommand{\Pixel}[1]{\mathrm{#1}}
\newcommand{\Frame}{\mathscr{F}}
\newcommand{\R}{\mathbb{R}}

\newcommand{\SE}[1]{\mathbb{SE}(#1)}
\newcommand{\rot}[1]{\rotatebox[origin=c]{90}{#1}}

\newcolumntype{P}[1]{>{\centering\arraybackslash}p{#1}}

\title{\LARGE \bf Robot Localization in Floor Plans\\ Using a Room Layout Edge
  Extraction Network}

\author{Federico Boniardi$^{*}$ \and Abhinav Valada$^{*}$ \and Rohit Mohan \and Tim Caselitz \and  Wolfram Burgard
  \thanks{$^{*}$These authors contributed equally to this work. All authors are with the Department of Computer Science, University of Freiburg, Germany. Wolfram Burgard is also with the Toyota Research Institute, Los Altos, USA.}%
}

\begin{document}

\onecolumn
{\Large

\noindent\textcopyright IEEE. Personal use of this material is permitted. Permission from IEEE must be obtained for all other uses, in any current or future media, including reprinting/republishing this material for advertising or promotional purposes, creating new collective works, for resale or redistribution to servers or lists, or reuse of any copyrighted component of this work in other works.\\

%
\noindent{Pre-print of the article that will appear in the\\ \textbf{Proceedings of the 2019 IEEE/RSJ International Conference on Intelligent Robots and Systems (IROS).}\\

\noindent{Please cite this paper as:}\\
F. Boniardi, A. Valada, R. Mohan, T. Caselitz, W. Burgard, "Robot Localization in Floor Plans Using a Room Layout Edge Extraction Network", \textit{Proceedings of the IEEE/RSJ International Conference on Intelligent Robots and Systems (IROS)}, 2019.\\

\noindent{BibTex:}\\
\\
@inproceedings$\lbrace$boniardi19iros,\\
author = $\lbrace$Federico Boniardi and Abhinav Valada and Rohit Mohan and Tim Caselitz and Wolfram Burgard$\rbrace$,\\
title = $\lbrace$Robot Localization in Floor Plans Using a Room Layout Edge Extraction Network$\rbrace$,\\
booktitle = $\lbrace$Proceedings of the IEEE/RSJ International Conference on Intelligent Robots and Systems (IROS)$\rbrace$,\\
year = $\lbrace$2019$\rbrace$,\\
$\rbrace$
}}
\twocolumn

\maketitle
\thispagestyle{empty}
\pagestyle{empty}

\begin{abstract}
  Indoor localization is one of the crucial enablers for deployment of service
  robots. Although several successful techniques for indoor localization have
  been proposed, the majority of them relies on maps generated from data
  gathered with the same sensor modality used for localization. Typically,
  tedious labor by experts is needed to acquire this data, thus limiting the
  readiness of the system as well as its ease of installation for inexperienced
  operators. In this paper, we propose a memory and computationally efficient
  monocular camera-based localization system that allows a robot to estimate its
  pose given an architectural floor plan. Our method employs a convolutional
  neural network to predict room layout edges from a single camera image and
  estimates the robot pose using a particle filter that matches the extracted
  edges to the given floor plan. We evaluate our localization system using
  multiple real-world experiments and demonstrate that it has the robustness and
  accuracy required for reliable indoor navigation.
\end{abstract}

\section{Introduction}
\label{sec:introduction}

Inexpensive sensors and ease of setup are widely considered as key enablers for
a broad diffusion of consumer-grade robotic applications. However, such
requirements pose technological challenges to manufacturers and developers due
to the limited quantity of sensory data and low quality of prior information
available to the robot. Particularly in the context of robot navigation, most of
the existing localization solutions require highly accurate maps that are built
upfront with the same sensor modality used for localizing the robot. Typically,
these maps are generated by collecting sensory measurements via teleoperation
and fusing them into a coherent representation of the environment using Simultaneous Localization and Mapping (SLAM) algorithms. Despite the
advances in the field, maps generated by SLAM systems can be affected by global
inconsistencies when perceptual aliasing or feature scarcity reduce the
effectiveness of loop closing approaches. In general, substantial expertise is
required to assess whether the quality of the generated maps is sufficient
for the planned deployment. For large-scale environments such as office
buildings, teleoperating the platform through the entire navigable area can be a
tedious and time-consuming operation. In order to address these issues, previous
works \cite{ito2014wrgbd, winterhalter15iros, boniardi17iros} have proposed to
leverage floor plans obtained from architectural drawings for accurate
localization as they provide a representation of the stable structures in the
environment. Furthermore, floor plans are often available from the blueprints
used for the construction of buildings. Alternatively, floor plans can also be
created with moderate effort using drawing utilities.

\begin{figure}
  \centering
  \begin{tabular}{@{}c@{}}
    \includegraphics[width=\columnwidth]{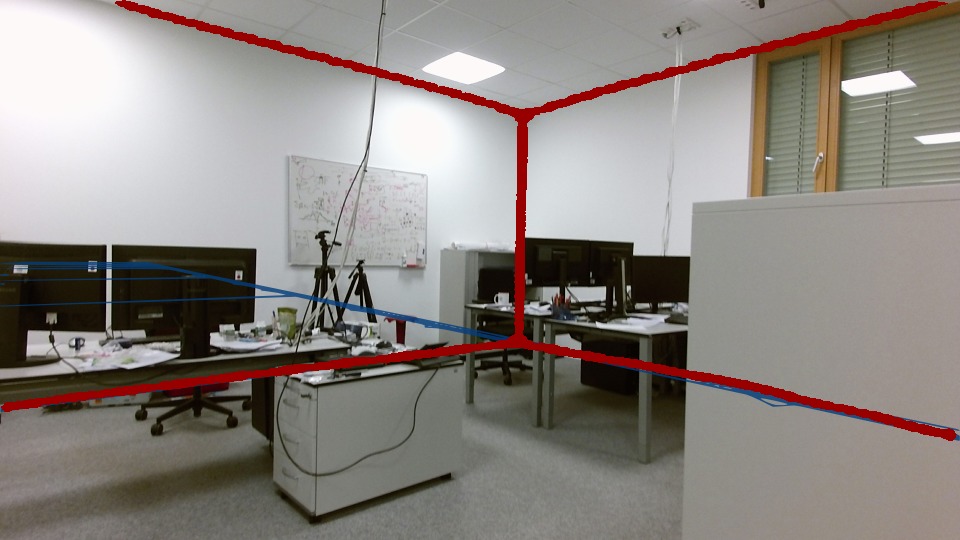} \\
    \includegraphics[width=\columnwidth]{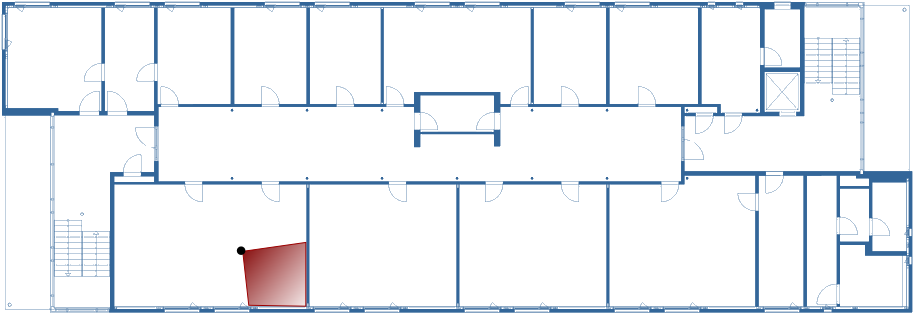} 
  \end{tabular}
      \vspace{-0.2cm}
  \caption{Our approach uses the proposed network to
    extract the room layout edges from an image (top) and compares it to a
    layout generated from a floor plan (bottom) to localize the robot. Our
    network is able to accurately predict layout edges even under severe
    occlusion, enabling robust robot localization.}
    \label{fig:cover-girl}
    \vspace{-0.5cm}
\end{figure}

Recently, computationally efficient approaches based on Convolutional Neural
Networks (CNNs) have been proposed for extracting structural information from
monocular images. This includes methods to extract room layout edges from
images~\cite{lin2018indoor,zhang2019edge}. However, these networks occasionally
predict discontinuous layout edges, even more in the presence of significant
clutter. In addition, room layouts can be inferred from floor plans under the
assumption that buildings consist only of orthogonal walls, also called
\emph{Manhattan world assumption}~\cite{coughlan2003manhattan}, and have
constant ceiling height.


Inspired by these factors, we propose a localization system that uses a
monocular camera and wheel odometry to estimate the robot pose using a given
floor plan. We propose a state-of-the-art CNN architecture to predict room layout edges
from a monocular image and apply a Monte Carlo Localization (MCL) method that compares
these edges with those inferred from a given floor plan. We evaluate
our proposed method in real-world scenarios, showing its robustness and accuracy
in challenging environments.

\section{Related Work}
\label{sec:related-work}


Several methods have been proposed to localize robots or, more generally,
devices, in 2D maps using RGB and range/depth measurements. For example, the
approaches proposed by Wolf~\etal \cite{wolf2002robust} and Bennewitz~\etal
\cite{bennewitz2006metric} use MCL and employ a database of images recorded in
an indoor environment. Mendez~\etal \cite{mendez2018sedar} proposed a sensor
model for MCL that leverages the semantics of the environment, namely doors,
walls and windows, obtained by processing RGB images with a CNN. They enhance
the standard \emph{likelihood fields} for the occupied space on the map with
suitable likelihood fields for doors and windows. Although such a sensor model
can be also adapted to handle range-less measurements, it shows increased
accuracy with respect to standard MCL only when depth measurements are
used. Winteralter~\etal \cite{winterhalter15iros} proposed a sensor model for
MCL to localize a Tango tablet in a floor plan. They extrude a full 3D model
from the floor plan and use depth measurements to estimate the current
pose. More recently, Lin~\etal \cite{lin2018floorplan} proposed a joint
estimation of the camera pose and the room layout using prior information from
floor plans. Given a set of partial views, they combine a floor plan extraction
method with a pose refinement process to estimate the camera poses.

The approaches described above rely on depth information or previously acquired
poses. Other methods only use monocular cameras to localize. Zhang and Kogadoga
\cite{zhang2005monocular} proposed a robot localization system based on wheel
odometry and monocular images. The system extracts edges from the image frame
and converts the floor edges into 2D world coordinates using the extrinsic
parameters of the camera. Such points are then used as virtual endpoints for
vanilla MCL. A similar approach by Unicomb~\etal \cite{unicomb2018monocular} was
proposed recently to localize a camera in a 2D map. The authors employ a CNN
for floor segmentation from which they identify which lines in an edge image
belong to the floor plan. The detected edges are reprojected into the 3D world
using the current estimate of the floor plane. They are then used as virtual
measurement in an \emph{extended Kalman filter}. Hile and
Boriello~\cite{hile2008positioning} proposed a system to localize a mobile phone
camera with respect to a floor plan by triangulating suitable features. They
employ RANSAC to estimate the relative 3D pose together with the feature
correspondences. Although the system achieves high accuracy, the features are
limited to corner points at the base of door frames and wall intersections. Therefore, the system is not usable outside corridors, due to occlusions and the
limited camera field-of-view. Chu~\etal \cite{chu2015you} use MCL to estimate
the 3D pose of a camera in an extruded floor plan. They proposed a sensor model
that incorporates information about the observed free-space, doors as well as
structural lines of the environment by leveraging a 3D metrical point cloud
obtained from monocular visual SLAM.

The method proposed in this work differs from the approaches above. Instead of
locally reconstructing the 3D world from camera observations and matching this
reconstruction to an extruded model of the floor plan, we project the lines
extracted from the floor plan into the camera frame. Our approach shares
similarity with the work of Chu and Chen \cite{chu2015towards}, Wang~\etal
\cite{wang2015lost} and Unicomb~\etal \cite{unicomb2018monocular}. In the first
two works the authors localize a camera using a 3D model extracted from a floor
plan. In order to score localization hypotheses, both systems use a
distance-transform-based cost function that encodes the misalignment on the
image plane between the structural lines extracted from the 3D model and the
edge image obtained by edge detection. In contrast to these approaches, we use a
CNN to reliably predict room layout edges in order to better cope with occlusion
due to clutter and furniture. Unicomb~\etal \cite{unicomb2018monocular} also
employ a CNN but they only learn to extract floor edges which is a limitation in
the case of clutter or occlusions. Furthermore, using a Kalman Filter approach
to project the measurement into the floor plan of the map can make the system
less robust to wrong initialization as the accuracy of the virtual measurement
is dependent on the current camera pose estimation. Finally, in contrast to
\cite{chu2015you} and \cite{wang2015lost}, we model the layout edges of the
floor plan from an image and wall corners without any prior 3D model.

Most of the CNN-based approaches for estimating room layout edges employ a
encoder-decoder topology with a standard classification network for the encoder
and utilize a series of deconvolutional layers for upsampling the feature maps
\cite{lin2018indoor,zhang2019edge,ren2016coarse,unicomb2018monocular}. Ren~\etal
\cite{ren2016coarse} proposed an architecture that employs the VGG-16 network
for the encoder followed by fully-connected layers and deconvolutional layers
that upsample to one quarter of the input resolution. The use of fully-connected
layers enables their network to have a large receptive field but at the cost of
loosing the feature localization ability. Lin~\etal~\cite{lin2018indoor}
introduced a similar approach with the stronger ResNet-101 backbone and model
the network in a fully-convolutional manner. Most recently, Zhang~\etal
\cite{zhang2019edge} proposed an architecture based on the VGG-16 backbone for
simultaneously estimating the layout edges as well as predicting the semantic
segmentation of the walls, floor and ceiling. In contrast to these networks, we
employ a more parameter efficient encoder with dilated convolutions and
incorporate the novel eASPP~\cite{valada2018self} for capturing large context,
complemented with an iterative training strategy that enables our network to
predict thin layout edges without discontinuities.


\section{Proposed Method}
\label{sec:proposed-method}

\begin{figure*}
\footnotesize
\centering
\includegraphics[width=\linewidth]{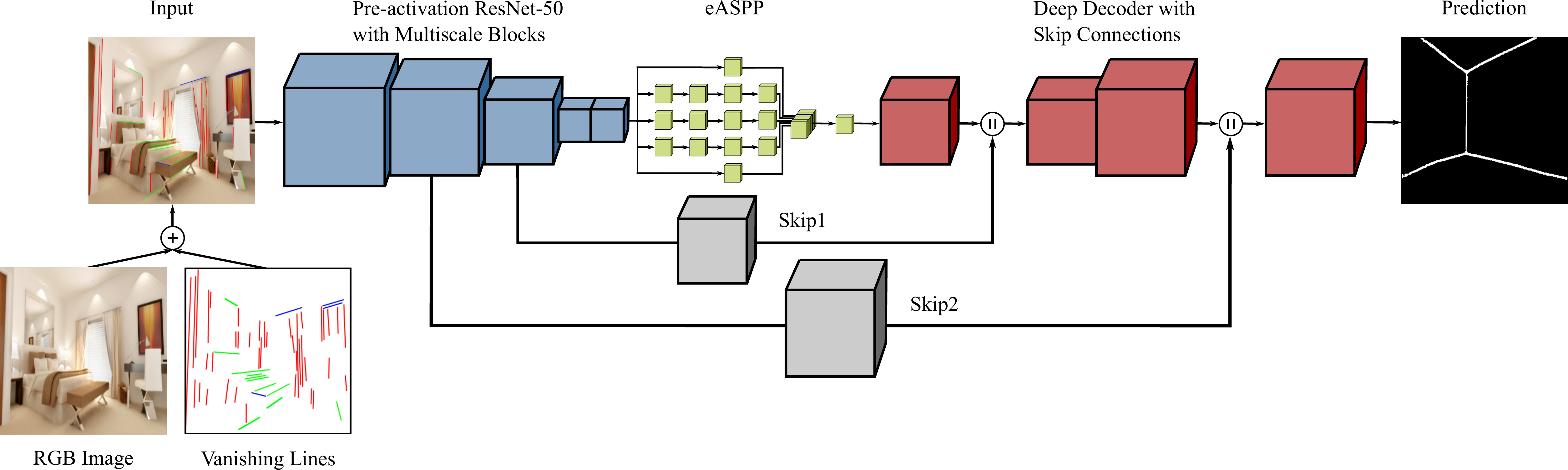}
\caption{Topology of our proposed architecture for extracting room layout edges
  that builds upon our AdapNet++ model~\cite{valada2018self}. The network takes
  colorized vanishing lines overlaid on the monocular image as input and
  reliably predicts the room layout edges.}
\label{fig:adapnetpp}
\vspace{-0.4cm}
\end{figure*}

In order to localize the robot in floor plans, we employ
MCL~\cite{thrun05probabilistic} with adaptive sampling. MCL applies Bayesian
recursive update
\begin{eqnarray}
  \label{eq:particle-filter}
    & \lefteqn{\Bel(\Pose{x}_{t})} \nonumber\\ & \propto & p(\Measurements_{t}
  \mid \Pose{x}_{t}) \int_{X} p(\Pose{x}_{t} \mid \Pose{x}_{t - 1},
  \Pose{u}_{t}) \Bel(\Pose{x}_{t-1}) d\Pose{x}_{t-1}
\end{eqnarray}
to a set of weighed hypothesis (particles) for the posterior distribution
$\Bel(\Pose{x}_{t}) \triangleq p(\Pose{x}_{t} \mid \Measurements_{1:t},
\Pose{u}_{1:t})$ of the robot pose $\Pose{x}_{t} \in \SE{2}$, given a sequence
of motion priors $\Pose{u}_{1:t}$ and sensor measurements
$\Measurements_{1:t}$. Whereas a natural choice for the proposal distribution
$p(\Pose{x}_{t} \mid \Pose{x}_{t - 1}, \Pose{u}_{t})$ is to apply the odometry
motion model with Gaussian noise, a suitable measurement model
$p(\Measurements_{t} \mid \Pose{x}_{t})$ based on the floor plan layout edges
has to be used, which we outline in the reminder of this section. To resample
the particle set, we use KLD-sampling, which is a well known sampling technique
that adapts the number of particles according to the Kullback-Leibler divergence
of the estimated belief and is an approximation of the true posterior distribution
\cite{fox2002kld}.

Note that in this work, we are only interested in the pose tracking problem,
that is, at every time $t > 0$ we estimate $bel(\Pose{x}_t \mid \Pose{x}_{0})$
given an initial coarse estimate $\Pose{x}_{0} \in \SE{2}$ of the starting
location of the robot. For real-world applications, solving the global
localization problem often not required as users can usually provide an initial
guess for the starting pose of the robot.

\subsection{Room Layout Edge Extraction Network}
\label{subsec:layout-extraction-from-images}

Our approach to estimate the room layout edges consists of two steps. In the
first step, we estimate the vanishing lines in a monocular image of the scene
using the approach of Hedau~\etal \cite{hedau2009recovering}. Briefly, we detect
vanishing lines by extracting line segments and estimating three mutually
orthogonal vanishing directions. Subsequently, we color the detected line
segments according to the vanishing point using a voting scheme. In the second
step, we overlay the estimated colorized vanishing lines on the monocular image
which is then input to our network for feature learning and
prediction. Utilizing the vanishing lines enables us to encode prior knowledge
about the orientation of the surfaces in the scene which accelerates the
training of the network and improves the performance in highly cluttered scenes.

\begin{figure}
\footnotesize
\centering
\setlength{\tabcolsep}{0.1cm}
\renewcommand{\arraystretch}{0.8}
\begin{tabular}{@{}P{0.2cm}P{4cm}P{4cm}@{}}
\rot{(a) ResNet-50} & \raisebox{-0.45\height}{\includegraphics[width=\linewidth]{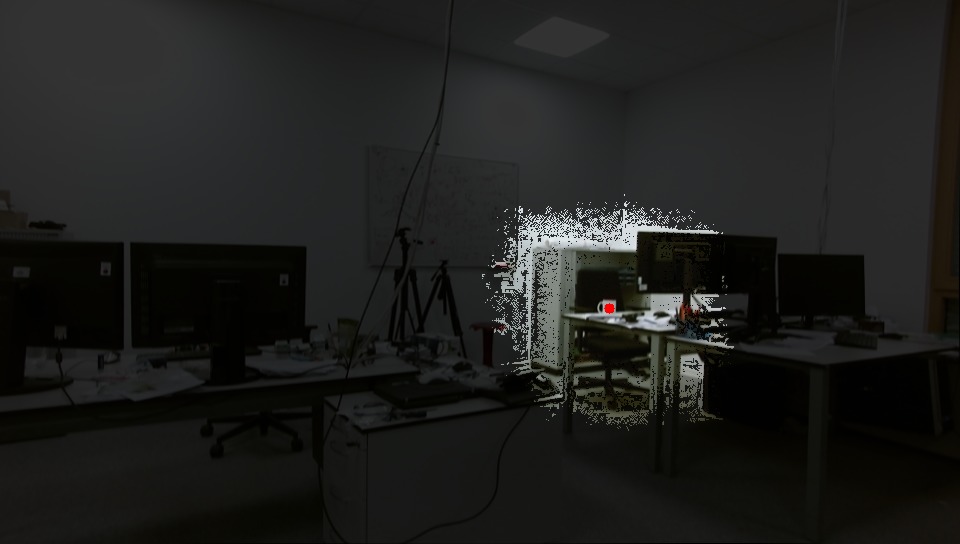}} & \raisebox{-0.45\height}{\includegraphics[width=\linewidth]{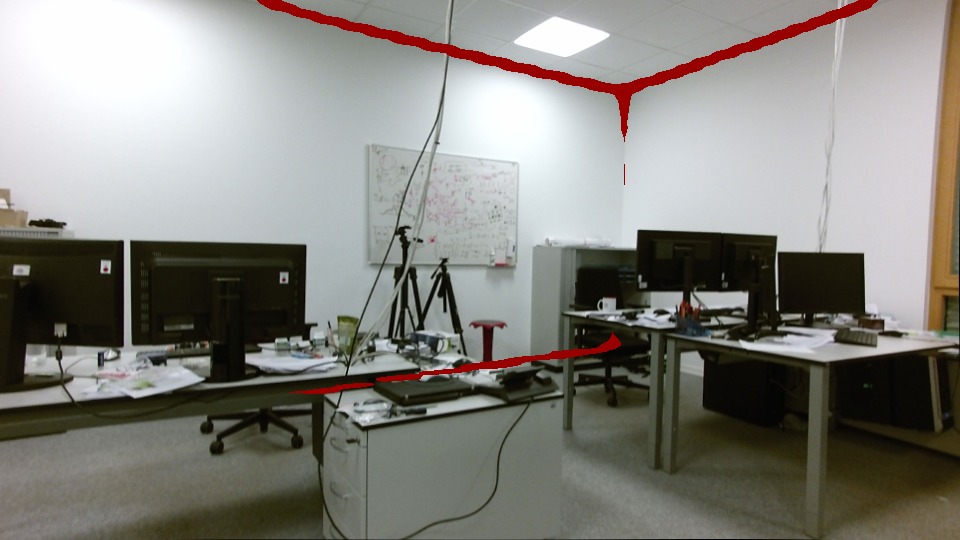}} \\
\\
\rot{(b) Ours} & \raisebox{-0.45\height}{\includegraphics[width=\linewidth]{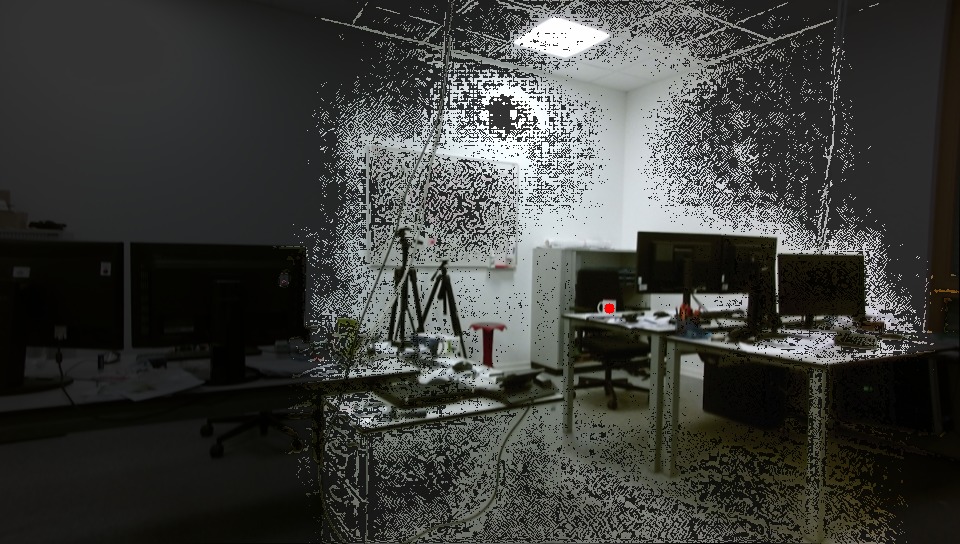}} & \raisebox{-0.45\height}{\includegraphics[width=\linewidth]{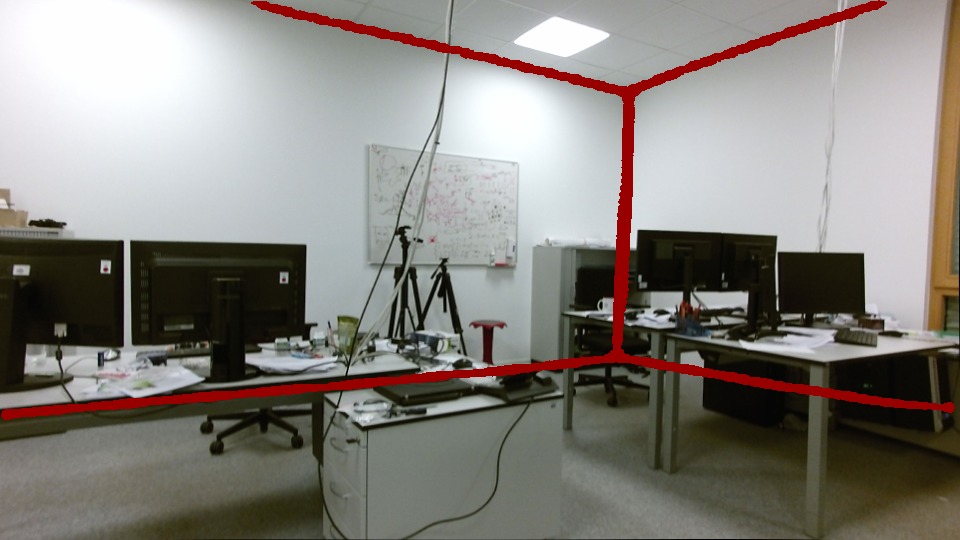}} \\
\\
& \raisebox{-0.45\height}{Receptive Field} & \raisebox{-0.45\height}{Prediction Overlay}
\end{tabular}
\caption{Comparison of the receptive field at the end of the encoder for the
  pixel annotated by the red dot (left). Our network has a much larger
  effective receptive field than the standard ResNet-50 model, which enables us
  to capture large context. This results in a more exhaustive description of the
  room layout edges (right).}
\label{fig:rescep}
\vspace{-0.2cm}
\end{figure}

The topology of our proposed architecture for learning to predict room layout
edges is shown in Figure~\ref{fig:adapnetpp}. We build upon our recently
introduced AdapNet++ architecture~\cite{valada2018self} which has four main
components. It consists of an encoder based on the full pre-activation ResNet-50
architecture~\cite{he2016identity} in which the standard residual units are
replaced with multiscale residual units~\cite{valada2017adapnet} encompassing parallel atrous convolutions with different dilation rates. We add dropout on the last two residual units to
prevent overfitting. The output of the encoder, which is 16-times downsampled
with respect to the input image, is then fed into the eASPP module. The eASPP
module has cascaded and parallel atrous convolutions to capture long-range
contexts with very large effective receptive fields. Having large effective
receptive fields is critical for estimating room layout edges as indoor scenes
are often significantly cluttered and the network needs to be able to capture
large contexts beyond the occluded regions. In order to illustrate this aspect, we compare the
empirical receptive field at the end of the eASPP of our network and the
receptive field at the end of the full pre-activation ResNet-50 architecture in
Figure~\ref{fig:rescep}. As we observe the receptive field for the pixel
annotated by the red dot, we see that the receptive field at the end of the
ResNet-50 architecture is not able to capture context beyond the clutter that
causes occlusion, whereas the larger receptive field of our network allows to
accurately predict the room layout edges even in the presence of severe occlusion.

In order to upsample the output of the eASPP back to the input image resolution,
we employ a decoder with three upsampling stages. Each stage employs a
deconvolution layer that upsamples the feature maps by a factor of two, followed
by two $3\times3$ convolution layers. We also fuse high-resolution encoder
features into the decoder to obtain smoother edges. We use the parameter
configuration for all the layers in our network as defined in the AdapNet++
architecture~\cite{valada2018self}, except for the last deconvolution layer in
which we set the number of filter channels to one and add a sigmoid activation
function to yield the room layout edges, which is thresholded to yield a binary
edge mask. We detail the training protocol that we employ in
Section~\ref{subsec:training}.

\subsection{Floor Plan Layout Edge Extraction}
\label{subsec:floorplan-layout-model}

As in our previous work~\cite{boniardi19ras}, we assume the floor plan to be
encoded as binary image $\mathcal{I} \in \{\textsc{o}, \textsc{f}\}^{H \times
  W}$ with resolution $\sigma$, a reference frame $\Frame \in \SE{2}$ and a set
of corner points $C \subset \R^{2}$ associated to some corner pixels and
expressed with respect to that reference frame. Corners are extracted by
preprocessing the map using standard corner detection algorithms and clustering
the resulting corners according to the relative distances in order to remove
duplicates. We embed the above structure in the 3D world and assume the above
entities to be defined in 3D while using the same notation.

\begin{figure}[t!]
  \centering
  \frame{\includegraphics[width=0.98\linewidth]{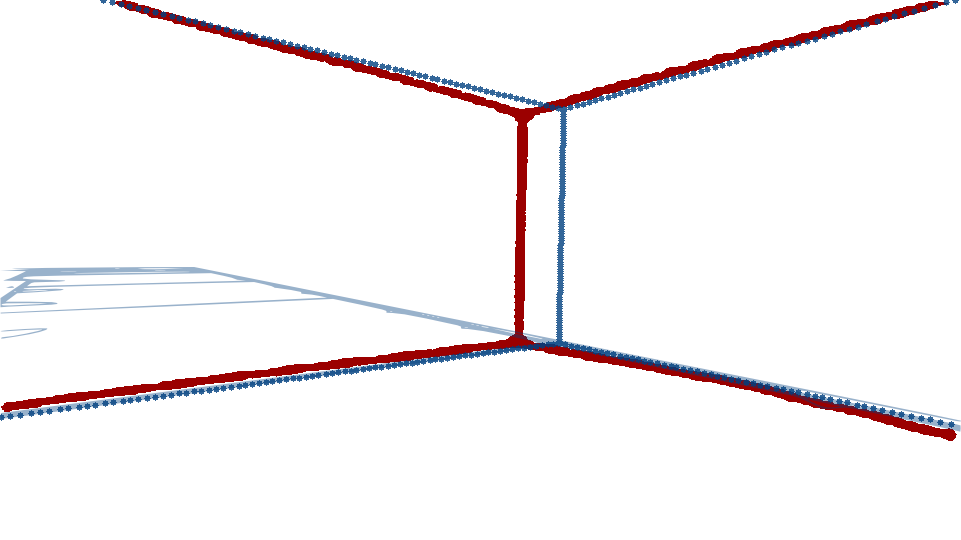}}
  \caption{\label{fig:floorplan-model} Example of layout $O_\Pose{x}$ extracted
    from the floor plan (blue) from a pose hypothesis $\Pose{x}$. The floor
    points are obtained via ray-casting, the ceiling lines by projecting up the
    floor points and the wall edge is obtained from the corner point. The
    floorplan (light blue) is overlayed for illustration purposes. The
    measurement model compares $O_\Pose{x}$ with the edge mask (red).}
\end{figure}

Similarly to Lin~\etal \cite{lin2018floorplan}, given a pose $\Pose{x} \in
\SE{3}$ on the floor plan and the extrinsic calibration parameters for the
optical frame of the camera ${}^{\Pose{r}}\Pose{T}_{\Pose{c}} \in \SE{3}$, we
can estimate the orthogonal projection of the camera's frustum onto the floor
plan plane (see Figure~\ref{fig:cover-girl}). Observe that such projection
defines two half-lines $\ell_{-} \triangleq \langle
[\Pose{x}\oplus{}^{\Pose{r}}\Pose{T}_{\Pose{c}}]_{xy}, \theta_{-} \rangle$ and
$\ell_{+} \triangleq \langle
[\Pose{x}\oplus{}^{\Pose{r}}\Pose{T}_{\Pose{c}}]_{xy}, \theta_{+} \rangle$,
where $[\Pose{x}\oplus{}^{\Pose{r}}\Pose{T}_{\Pose{c}}]_{xy} \in \R^{2}$ is the
orthogonal projection of the origin of the optical frame onto the floor plan
(origin of the half-lines), and $\theta_{\pm} \in (-\pi,\pi]$ are the ray
  directions with respect to the 2D reference frame $\Frame$.  Such rays define
  an angular range $[\theta_{-}, \theta_{+}] \subset (-\pi,\pi]$ that
    approximates the planar field-of-view (FoV) of the camera. Accordingly, we
    can approximate the layout room edges of the visible portion of the floor
    plan with a discrete set of points $O_{\Pose{x}} \subset \R^{3}$ estimated
    or extruded from the floor plan image. More specifically, we construct
    $O_{\Pose{x}}$ by inserting the points obtained by ray-casting within the
    camera FoV as well as their counterparts on the ceiling, obtained by
    elevating the ray-casted points by the height of the building, which we
    assume to be known upfront. Moreover, to complete the visible layout edges,
    we add to $O_\Pose{x}$ those corners in $C$ whose lines of sight from
    $[\Pose{x}\oplus{}^{\Pose{r}}\Pose{T}_{\Pose{c}}]_{xy}$ fall within the 2D
    FoV of the camera together with their related ceiling points as well as the
    set of intermediate points sampled along the connecting vertical line (see
    Figure~\ref{fig:floorplan-model}). The visibility of each corner point can
    be inferred, again, by ray-casting along the direction of each line of
    sight. Observe that, although ray-casting might be computationally expensive
    due to a high resolution $\sigma$, speed up can be achieved by ray-casting
    on floor plan images with a lower resolution.

\subsection{Measurement Model}
\label{subsec:layout-measurement-model}

Given an input image and the related layout edge mask $\Measurements$, we define
the observation model of each pose hypotheses as follows: for any pose $\Pose{x}
\in \SE{2}$ on the floor plan, we set
\begin{equation}
  \label{eq:sensor-model}
  \log p(\Measurements \mid \Pose{x}) =
  -\frac{1}{2|O_{\Pose{x}}|\sigma_{z}^{2}} \sum_{o \in O_{\Pose{x}}}
  \min\left\{d(\pi(o), \Measurements), \delta \right\}^{2}
\end{equation}
where $\delta > 0$ (in pixel) is a saturation term used to avoid excessive
down-weighing of particles whenever a measurement cannot be explained by the
floor plan model, $\sigma_{\Measurement} > 0$ (in pixel) is a tolerance term
that encodes the expected pixel noise in the layout edge mask $\Measurements$,
$\pi(o)$ is the perspective camera transformation that projects 3D world points
into the image plane, and $d(\Pixel{p}, \Measurements)$ is the distance of pixel
$\Pixel{p}$ to the closest pixel in edge layout mask.

\begin{figure*}
  \centering
  \begin{tabular}{@{}c@{}}
    \includegraphics[width=0.86\columnwidth]{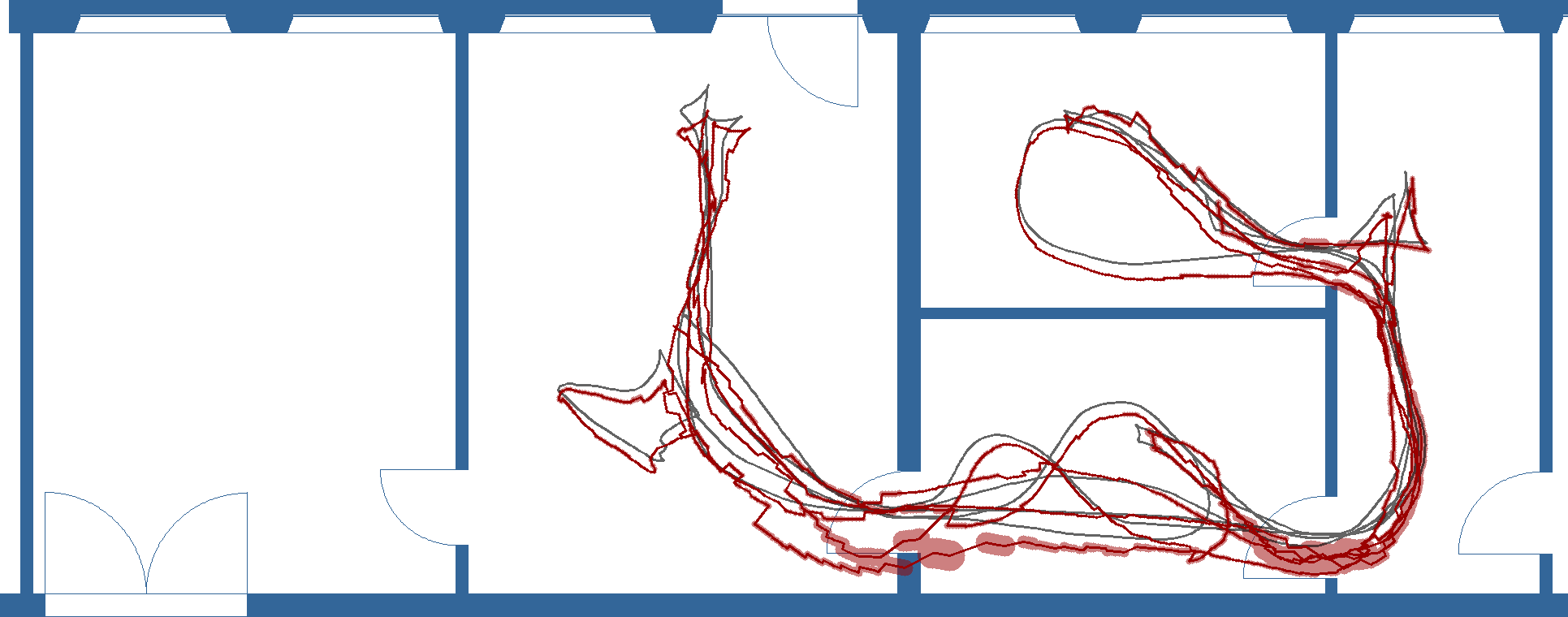} \hspace{1cm}
    \includegraphics[width=\columnwidth]{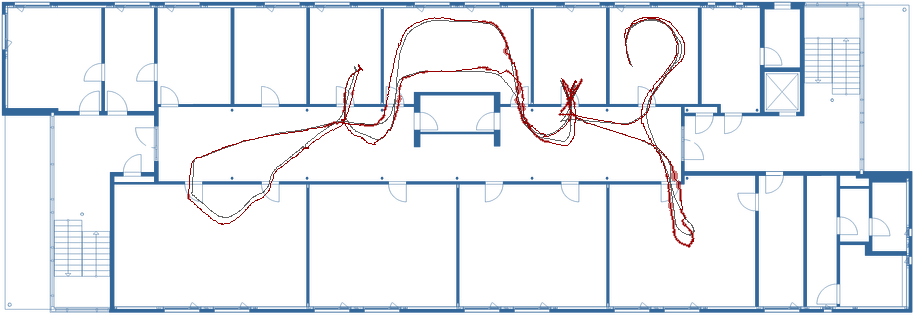} \\
    \includegraphics{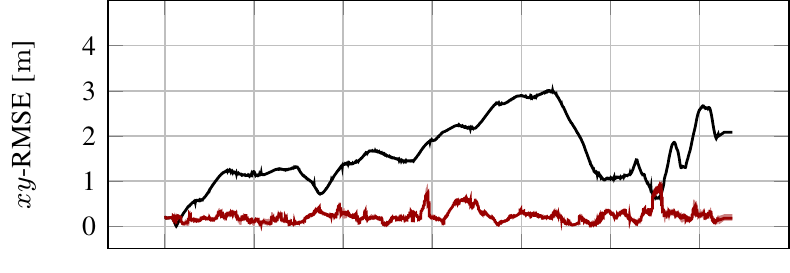} \hfill    
    \includegraphics{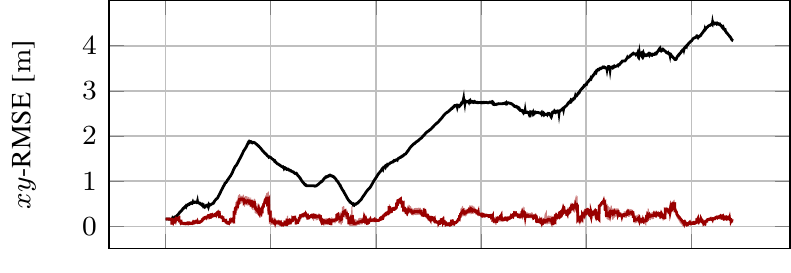} \\
    \includegraphics{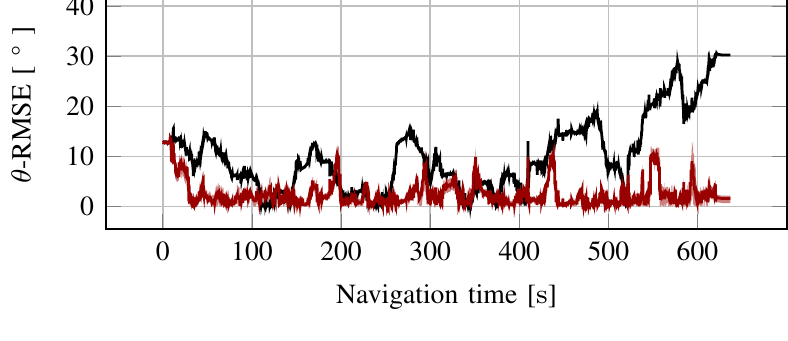} \hfill
    \includegraphics{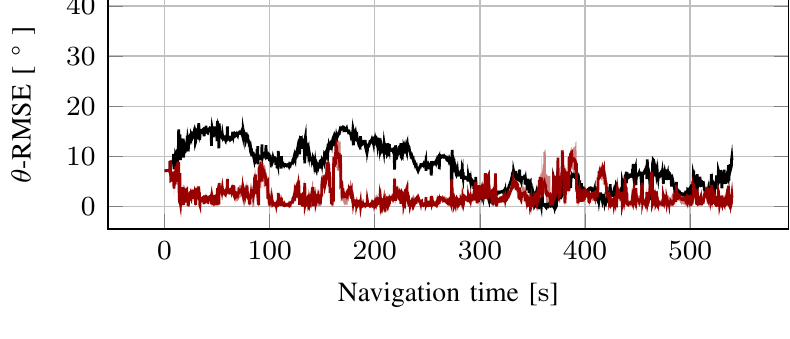}
  \end{tabular}
  \caption{\label{fig:experiments} Accuracy of the localization system for
    \emph{Fr078-1} (left) and \emph{Fr080} (right). Top: the estimated mean
    trajectory (red) compared to the approximate ground-truth (gray). The red
    shadowed area represents the translational standard deviation of each pose
    estimate. Middle and bottom: The linear and angular RMSE (red) and the error
    produced by drifting odometry (black). The The red areas delimit the errors
    for the worst and best pose estimation.}
    \vspace{-0.4cm}
\end{figure*}

\section{Experimental Evaluation}
\label{sec:experiments}

To evaluate the performance of our proposed method, we recorded three datasets
in two buildings of the University of Freiburg (building 078 and 080). We will
henceforth refer to them as \emph{Fr078-1} (\SI{113}{\m} long), \emph{Fr078-2}
(\SI{179}{\m} long) and \emph{Fr080} (\SI{108}{\m} long). \emph{Fr078-1} and
\emph{Fr078-2} aim to emulate an apartment-like structure while \emph{Fr080} was
obtained obtained in a standard office building. For all the experiments, we
used a Festo Robotino omnidirectional platform and the RGB images obtained from
a Microsoft Kinect~V2 mounted on board the robot. The robot moved with an
average speed of approximately $\SI{0.2}{\m/\s}$ and $\SI{15}{\degree/\s}$ and
maximum of $\SI{0.5}{\m/\s}$ and $\SI{50}{\degree/\s}$. Since no ground-truth
was available for these experiments, we employed the localization system
proposed in \cite{boniardi19ras} using an Hokuyo UTM-30LX laser rangefinder also
mounted on the robot to provide a reference trajectory for the evaluation. Since
the trajectories estimated by \cite{boniardi19ras} are highly accurate, we will
henceforth consider them to be the (approximate) ground-truth. For each dataset,
we run $\num{25}$ experiments to account for the randomness of MCL and consider
the estimated pose at each time to be the average pose over the runs.

In addition, we benchmark the performance of our room layout edge estimation
network on the challenging LSUN Room Layout Estimation
dataset~\cite{zhang2015large} consisting of 4,000 images for training, 394
images for validation and 1,000 images for testing. We employ augmentation
strategies such as horizontal flipping, cropping and color jittering to increase
the number of training samples. We report results in terms of edge error, which
can be computed as the Euclidean distance between the estimated layout edges and
the ground-truth edge map, normalized by the number of pixel in each mask. In
order to facilitate comparison with previous approaches~\cite{zhang2019edge}, we
also report the fixed contour threshold (ODS) and the per-image best threshold
(OIS)~\cite{arbelaez2011contour} metrics.

\subsection{Implementation}
\label{subsec:implementation}

In all the experiments we used the same set of parameters. To extract the room
layout of the floor plans we removed single pixel lines as well as close any
doors gaps and narrow passages by using an erosion/dilation and dilation/erosion
pass respectively on floor plans with resolution $\sigma =
\SI{1}{\cm\per\px}$. Similarly, the \emph{Harris corner detector} implementation
of OpenCV was utilized to extract the corner pixels on the floor plan image. In
our implementation of MCL we set $\sigma_{\Measurement} = \SI{10}{\px}$ and
$\delta = \SI{25}{\px}$. To compute the predicted layout $O_{\Pose{x}}$, we
subsampled the 2D camera FoV with $\num{150}$ rays and approximated the vertical
edges of the layout with $\num{100}$ points. Localization updates occurred
whenever the motion prior from wheel odometry reported a linear or angular
relative motion exceeding $\SI{25}{\cm}$ or approximately $\SI{15}{\degree}$
($\SI{0.25}{\rad}$) respectively and used 1,500 and 5,000 as minimum/maximum
number of particles to approximate the robot belief.


\subsection{Network Training}
\label{subsec:training}

We used the TensorFlow deep learning library for the network implementation and
we trained our model on images resized to a resolution of $320\times 320$
pixels.  The output of our network has the same resolution as the input
image. To generate the ground-truth data for training, we first converted the
LSUN room layout ground-truth to a binary edge map where the edge lines have a
width of 6 pixels. We dilated the edges with a $5\times5$ kernel for $d_e$
number of iterations, where $e$ is the number of epochs for which we trained
using this dilation factor. We then applied Gaussian blur with a kernel of
$21\times 21$ pixels and $\sigma = 6$ for smoothing the edge boundaries. We
employed a four stage training procedure and began training with the
ground-truth edges dilated with $d_{6} = 5$ and in subsequent stages reduced the
amount of edge dilation to $d_{14} = 3$, $d_{20} = 1$ and $d_{26} =
0$. Intuitively this process can be described as starting the training with
thick layout edges and gradually thinning the edge thickness as the training
progresses. Employing this gradual thinning approach improves convergence and enables the network to predict precise thin edges, as opposed to training only
with a fixed edge width. Lin~\etal \cite{lin2018indoor} employ a similar
training strategy that adaptively changes the edge thickness according to the
gradient, however our training strategy resulted in a better performance.

We used the He initialization~\cite{he2015delving} for all the layers of our
network and the cross-entropy loss function for training. For optimization, we
used Adam solver with $\beta_1=0.9$, $\beta_2=0.999$ and
$\epsilon=10^{-10}$. Additionally, we suppressed the gradients of non-edge
pixels by multiplying them with a factor of 0.2 in order to prevent the network
from converging to zero, which often occurs due to the imbalance between edge
and non-edge pixels. We trained our model for a total of 66 epochs with an
initial learning rate of $\lambda_0=10^{-4}$ and a mini-batch size of 16, which
takes about 18 hours on an NVIDIA TITAN X GPU.

\subsection{Evaluation of Layout Edge Estimation}
\label{subsec:outline-detection-accuracy}

In order to empirically evaluate the performance of our room layout edge
extraction network, we performed evaluations on the LSUN benchmark in comparison
to state-of-the-art approaches~\cite{zhang2019edge, lin2018indoor,
  he2016identity}. The results are reported in
Table~\ref{tab:benchmarkingEdge}. Our network achieved an edge error of
$8.33$ which corresponds to an improvement of $2.91$ over the
previous state-of-the-art. We also observe higher ODS as well as OIS scores,
thereby setting the new state-of-the-art on the LSUN benchmark for room layout
edge estimation. The improvement achieved by our network can be attributed to
its large effective receptive field, which enables it to capture more global
context. Moreover, our iterative training strategy allows for estimation of thin
layout edges without significant discontinuities.

\begin{figure} 
\footnotesize
\centering
\setlength{\tabcolsep}{0.1cm}
\renewcommand{\arraystretch}{0.5}
\begin{tabular}{@{}P{2cm}P{2cm}P{2cm}P{2cm}@{}}
Input Image & Lin~\etal \cite{lin2018indoor} & Zhang~\etal \cite{zhang2019edge} & Ours \\
\\
\includegraphics[width=\linewidth]{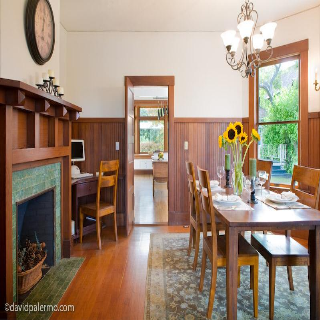} & \includegraphics[width=\linewidth]{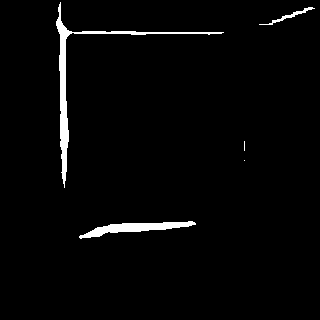} & \includegraphics[width=\linewidth]{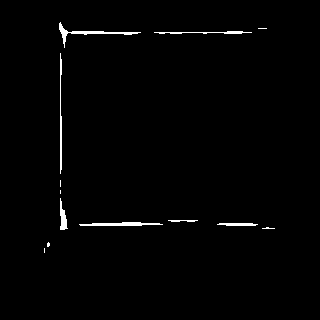} & \includegraphics[width=\linewidth]{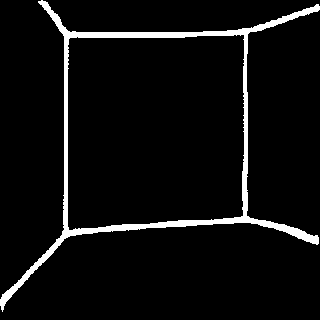} \\
\\
\includegraphics[width=\linewidth]{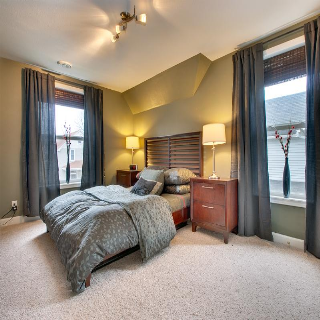} & \includegraphics[width=\linewidth]{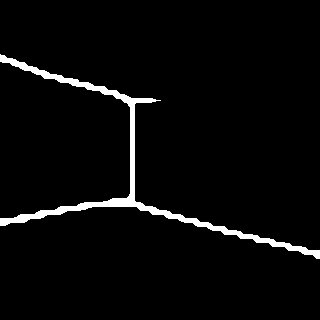} & \includegraphics[width=\linewidth]{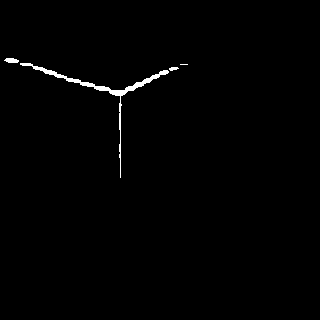} & \includegraphics[width=\linewidth]{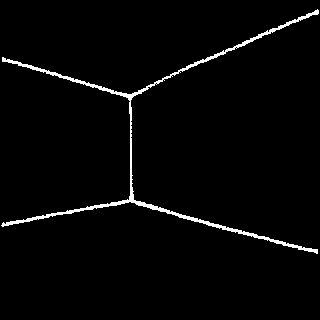} \\
\\
\includegraphics[width=\linewidth]{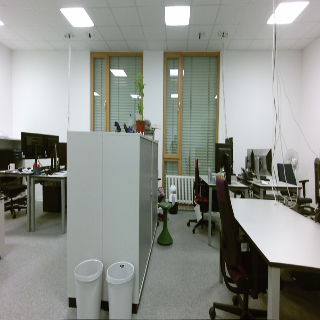} & \includegraphics[width=\linewidth]{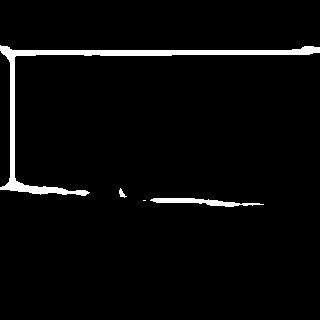} & \includegraphics[width=\linewidth]{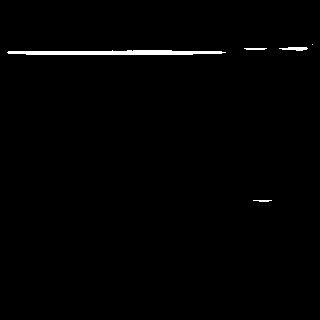} & \includegraphics[width=\linewidth]{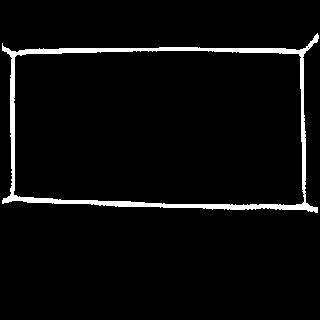} \\
\\
\includegraphics[width=\linewidth]{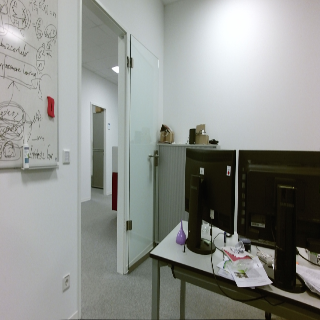} & \includegraphics[width=\linewidth]{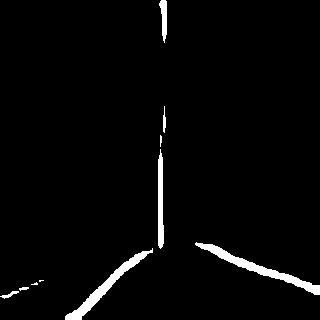} & \includegraphics[width=\linewidth]{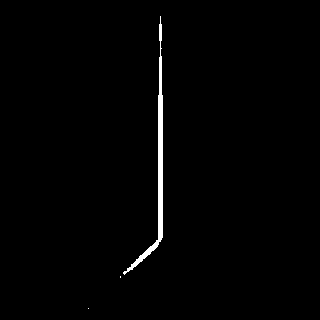} & \includegraphics[width=\linewidth]{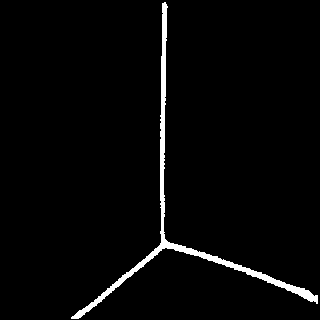} \\
\end{tabular}
\caption{Qualitative room layout edge estimation results on the LSUN validation
  set (first two rows) and on \emph{Fr080} (last two rows). Compared to the other
  methods, our network reliably predicts continuous layout edges even under substantial
  occlusion.}
\label{fig:edgeQual}
\end{figure}

Qualitative comparisons of room layout edge estimation are reported in
Figure~\ref{fig:edgeQual}. The first two and last two rows show prediction
results on the LSUN validation set and \emph{Fr080} dataset respectively. Note that we
only trained our network on the LSUN training set. We can see that the previous
state-of-the-art networks are less effective in predicting the layout edges in
the presence of large objects in the scene that cause significant occlusions,
whereas our network is able to leverage its large receptive field to more
reliably capture the layout edges. We can also observe that the prediction of
the other networks are more irregular and sometimes either too thin, thus
resulting in discontinuous layouts, or too thick, reducing the effectiveness of
the sensor model described in Section~\ref{subsec:layout-measurement-model}. In
these scenarios our network is able to accurately predict the layout edges
without discontinuities and generalize effectively to previously unseen
environments.

\subsection{Ablation Study of Layout Edge Estimation Network}
\label{subsec:outline-detection-ablation}

\begin{table}
\centering
\caption{Benchmarking edge layout estimation on the LSUN dataset.}
\label{tab:benchmarkingEdge}
\begin{tabular}{@{}lcccr@{}}
\toprule
\textbf{Network} & \textbf{Edge Error} & \textbf{ODS} & \textbf{OIS} & \textbf{Parameters} \\
\midrule
ResNet50-FCN~\cite{he2016identity} & 18.36 & 0.213 & 0.227 & 23.57\,M \\
Lin~\etal \cite{lin2018indoor} & 10.72 & 0.279 & 0.284 & 42.29\,M \\
Zhang~\etal \cite{zhang2019edge} & 11.24 & 0.257 & 0.263 & 138.24\,M \\
\midrule
Ours & \textbf{8.33} & \textbf{0.310} & \textbf{0.316} & 30.19\,M\\
\bottomrule
\end{tabular}
\end{table}

\begin{table}
\centering
\caption{Layout edge estimation network configuration.}
\label{tab:ablationEdge}
\begin{tabular}{@{}ccccr@{}}
\toprule
\textbf{Model} & \textbf{Output} & \textbf{Background} & \textbf{Vanishing} & \textbf{Edge} \\
 & \textbf{Resolution} & \textbf{Weight} & \textbf{Lines} & \textbf{Error} \\
\midrule
M1 & 1/4 & 0.0 & - & 10.99 \\
M2 & 1/4 & 0.2 & - & 10.61 \\
M3 & 1/2 & 0.2 & - & 10.13 \\
M4 & Full & 0.2 & - & 9.46 \\
M5 & Full & 0.2 & \checkmark & 8.33 \\
\bottomrule
\end{tabular}
\end{table}

We evaluated the performance of our network through the different stages of the
upsampling. Referring to Table~\ref{tab:ablationEdge}, the M1 model upsamples
the eASPP output to one quarter the resolution of the input image and achieves
an edge error of $10.99$. In the subsequent M2 and M3 models, we suppress the
gradients of the non-edge pixels with a factor of 0.2 and upsample the eASPP
output to half the resolution of the input image, which reduces the edge error
by $0.86$. Finally, in the M4 model, we upsample back to the full input image
resolution and in the M5 model we overlay the colorized vanishing lines by
adding these channels to the RGB image. Our final M5 model achieves a reduction
of $2.66$ in the edge error compared to base AdapNet++ model.

\subsection{Localization Robustness and Accuracy}
\label{subsec:localization-accuracy-and-robustness}

In all the experiments, the robot was initialized within $\SI{10}{\cm}$ and
$\SI{15}{\degree}$ from the ground-truth pose. As shown in
Figure~\ref{fig:experiments}, the robot was always able to estimate its current
pose and non negligible errors were reported only in a specific situation. As
shown in Figure~\ref{fig:experiments}, in \emph{Fr078-1}, the robot failed
temporarily to track its current pose while traversing a doorway (scattered
trajectory at the bottom of the map). The error was due to the camera image
capturing both the next room (predominant view) and the current room (limited
view). This caused the network to only predict the layout edges for largest room
view, before having entered the next room. The linear RMSE of the mean
trajectory reached approximately $\SI{1}{\m}$. Nonetheless, the robot was
subsequently able to localize itself with an accuracy similar to the average
accuracy over the entire experiment whenever new observations were collected.

Overall, the proposed method delivered an average linear RMSE of $\SI{227 +-
  137}{\mm}$ and $\SI{245 +- 137}{\mm}$ as well as an average angular RMSE of
  $\SI{2.5 +- 2.5}{\degree}$ and $\SI{2.5 +- 2.3}{\degree}$ in \emph{Fr078-1}
  and \emph{Fr078-2} respectively. Similar results were recorded for
  \emph{Fr080}, with an average linear and angular RMSE of $\SI{223 +-
    126}{\mm}$ and $\SI{2.3 +- 2.0}{\degree}$ respectively.

\subsection{Runtime}
\label{subsec:runtime}

\begin{figure}
  \centering
  \begin{tabular}{@{}c@{}}
    \includegraphics{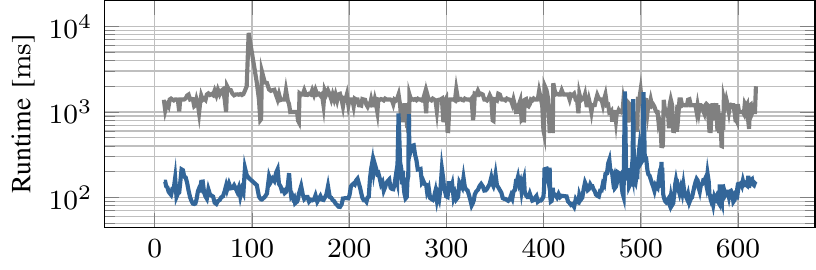} \\
    \includegraphics{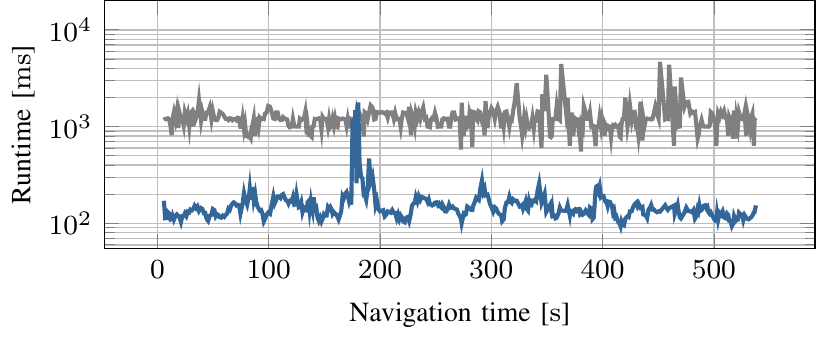}
  \end{tabular}
  \caption{\label{fig:runtime} Total runtime in milliseconds (blue line) for the
    network processing and the localization update in \emph{Fr078-1} (top) and
    \emph{Fr080} (bottom). The gray line represents the available runtime
    between two consecutive localization updates. Each runtime is computed as
    average runtime over the runs.}
\end{figure}

We used a 8-core $\SI[mode=text]{4.0}{\GHz}$ Intel~Core~i7 CPU and a NVIDIA
GeForce~980M GPU in all experiments. On average, the system required $\SI{30 +-
  14}{\ms}$ for the MCL update, while the inference time for the proposed
network was $\SI{39}{\ms}$. In addition, the Manhattan line extraction took
$\SI{80 +- 116}{\ms}$. The high peaks in Figure~\ref{fig:runtime} are due to the
Manhattan lines extraction step, which runs on an external Matlab script and
therefore it can be further optimized. As shown in Figure~\ref{fig:runtime}, our
proposed approach runs in real-time on consumer grade hardware.

\section{Conclusions}
\label{sec:conclusions}

In this work, we presented a robot localization system that uses wheel odometry
and images from a monocular camera to estimate the pose of a robot in a floor
plan. We utilize a novel convolutional neural network tailored to predict the room
layout edges and employ Monte Carlo Localization with a sensor model that scores
the overlap of the predicted layout edge mask and the expected layout edges
generated from a floor plan image. Experiments in complex real-world
environments demonstrate that our proposed system is able to robustly estimate
the pose of the robot even in challenging conditions such as severe
occlusion. In addition, our network for room layout edge estimation achieves
state-of-the-art performance on the challenging LSUN benchmark and generalize effectively to
previously unseen environments with complex room layouts.


\vspace{0.1cm}
\bibliographystyle{IEEEtran}
\bibliography{references}

\begin{thebibliography}{10}
\providecommand{\url}[1]{#1}
\csname url@samestyle\endcsname
\providecommand{\newblock}{\relax}
\providecommand{\bibinfo}[2]{#2}
\providecommand{\BIBentrySTDinterwordspacing}{\spaceskip=0pt\relax}
\providecommand{\BIBentryALTinterwordstretchfactor}{4}
\providecommand{\BIBentryALTinterwordspacing}{\spaceskip=\fontdimen2\font plus
\BIBentryALTinterwordstretchfactor\fontdimen3\font minus
  \fontdimen4\font\relax}
\providecommand{\BIBforeignlanguage}[2]{{%
\expandafter\ifx\csname l@#1\endcsname\relax
\typeout{** WARNING: IEEEtran.bst: No hyphenation pattern has been}%
\typeout{** loaded for the language `#1'. Using the pattern for}%
\typeout{** the default language instead.}%
\else
\language=\csname l@#1\endcsname
\fi
#2}}
\providecommand{\BIBdecl}{\relax}
\BIBdecl

\bibitem{ito2014wrgbd}
S.~Ito, F.~Endres, M.~Kuderer, G.~D. Tipaldi, C.~Stachniss, and W.~Burgard,
  ``{W-RGB-D}: floor-plan-based indoor global localization using a depth camera
  and wifi,'' in \emph{Proc.~of the IEEE International Conference on Robotics
  and Automation}, 2014.

\bibitem{winterhalter15iros}
W.~Winterhalter, F.~Fleckenstein, B.~Steder, L.~Spinello, and W.~Burgard,
  ``Accurate indoor localization for {RGB-D} smartphones and tablets given {2D}
  floor plans,'' in \emph{Proc.~of the IEEE/RSJ International Conference on
  Intelligent Robots and Systems}, 2015.

\bibitem{boniardi17iros}
F.~Boniardi, T.~Caselitz, R.~K{\"u}mmerle, and W.~Burgard, ``Robust
  {L}i{DAR}-based localization in architectural floor plans,'' in
  \emph{Proc.~of the IEEE/RSJ International Conference on Intelligent Robots
  and Systems}, 2017.

\bibitem{lin2018indoor}
H.~J. Lin, S.-W. Huang, S.-H. Lai, and C.-K. Chiang, ``Indoor scene layout
  estimation from a single image,'' in \emph{Proc.~of the IEEE International
  Conference on Pattern Recognition}, 2018.

\bibitem{zhang2019edge}
W.~Zhang, W.~Zhang, and J.~Gu, ``Edge-semantic learning strategy for layout
  estimation in indoor environment,'' \emph{arXiv preprint arXiv:1901.00621},
  2019.

\bibitem{coughlan2003manhattan}
J.~Coughlan and A.~Yuille, ``Manhattan world: Orientation and outlier detection
  by {B}ayesian inference,'' \emph{Neural Computation}, vol.~15, no.~5, pp.
  1063--1088, 2003.

\bibitem{wolf2002robust}
J.~Wolf, W.~Burgard, and H.~Burkhardt, ``Robust vision-based localization for
  mobile robots using an image retrieval system based on invariant features,''
  in \emph{Proc.~of the IEEE International Conference on Robotics and
  Automation}, 2002.

\bibitem{bennewitz2006metric}
M.~Bennewitz, C.~Stachniss, W.~Burgard, and S.~Behnke, ``Metric localization
  with scale-invariant visual features using a single perspective camera,'' in
  \emph{European Robotics Symposium}, 2006.

\bibitem{mendez2018sedar}
O.~Mendez, S.~Hadfield, N.~Pugeault, and R.~Bowden, ``{S}e{DAR}-semantic
  detection and ranging: Humans can localise without lidar, can robots?'' in
  \emph{Proc.~of the IEEE International Conference on Robotics and Automation},
  2018.

\bibitem{lin2018floorplan}
C.~Lin, C.~Li, Y.~Furukawa, and W.~Wang, ``Floorplan priors for joint camera
  pose and room layout estimation,'' \emph{arXiv preprint arXiv:1812.06677},
  2018.

\bibitem{zhang2005monocular}
Z.~Zhang and S.~Kodagoda, ``A monocular vision based localizer,'' in
  \emph{Proc.~of the Australasian Conference on Robotics and Automation}.\hskip
  1em plus 0.5em minus 0.4em\relax Australian Robotics and Automation
  Association, 2005.

\bibitem{unicomb2018monocular}
J.~Unicomb, R.~Ranasinghe, L.~Dantanarayana, and G.~Dissanayake, ``A monocular
  indoor localiser based on an extended kalman filter and edge images from a
  convolutional neural network,'' in \emph{Proc.~of the IEEE/RSJ International
  Conference on Intelligent Robots and Systems}, 2018.

\bibitem{hile2008positioning}
H.~Hile and G.~Borriello, ``Positioning and orientation in indoor environments
  using camera phones,'' \emph{Computer Graphics and Applications}, vol.~28,
  no.~4, 2008.

\bibitem{chu2015you}
H.~Chu, D.~Ki~Kim, and T.~Chen, ``You are here: Mimicking the human thinking
  process in reading floor-plans,'' in \emph{Proc.~of the IEEE International
  Conference on Computer Vision}, 2015.

\bibitem{chu2015towards}
Towards indoor localization with floorplan-assisted priors. \emph{Available
  online}. Accessed on January 2019.

\bibitem{wang2015lost}
S.~Wang, S.~Fidler, and R.~Urtasun, ``Lost shopping! monocular localization in
  large indoor spaces,'' in \emph{Proc.~of the IEEE International Conference on
  Computer Vision}, 2015.

\bibitem{ren2016coarse}
Y.~Ren, S.~Li, C.~Chen, and C.-C.~J. Kuo, ``A coarse-to-fine indoor layout
  estimation (cfile) method,'' in \emph{Asian Conference on Computer Vision},
  2016, pp. 36--51.

\bibitem{valada2018self}
A.~Valada, R.~Mohan, and W.~Burgard, ``Self-supervised model adaptation for
  multimodal semantic segmentation,'' \emph{arXiv preprint arXiv:1808.03833},
  2018.

\bibitem{thrun05probabilistic}
S.~Thrun, W.~Burgard, and D.~Fox, \emph{Probabilistic Robotics}.\hskip 1em plus
  0.5em minus 0.4em\relax MIT Press, 2005.

\bibitem{fox2002kld}
D.~Fox, ``{KLD}-sampling: Adaptive particle filters,'' in \emph{Advances in
  Neural Information Processing Systems}, 2002.

\bibitem{hedau2009recovering}
V.~Hedau, D.~Hoiem, and D.~Forsyth, ``Recovering the spatial layout of
  cluttered rooms,'' in \emph{Proc.~of the IEEE International Conference on
  Computer Vision}, 2009.

\bibitem{he2016identity}
K.~He, X.~Zhang, S.~Ren, and J.~Sun, ``Identity mappings in deep residual
  networks,'' in \emph{Proc.~of the European Conference on Computer Vision},
  2016.

\bibitem{valada2017adapnet}
A.~Valada, J.~Vertens, A.~Dhall, and W.~Burgard, ``Adapnet: Adaptive semantic
  segmentation in adverse environmental conditions,'' in \emph{Proc.~of the
  IEEE International Conference on Robotics and Automation}, 2017, pp.
  4644--4651.

\bibitem{boniardi19ras}
F.~Boniardi, T.~Caselitz, R.~K{\"u}mmerle, and W.~Burgard, ``A pose graph-based
  localization system for long-term navigation in {CAD} floor plans,''
  \emph{Robotics and Autonomous Systems}, vol. 112, pp. 84 -- 97, 2019.

\bibitem{zhang2015large}
Y.~Zhang, F.~Yu, S.~Song, P.~Xu, A.~Seff, and J.~Xiao. Large-scale scene
  understanding challenge: Room layout estimation. \emph{Available online}.
  Accessed on January 2019.

\bibitem{arbelaez2011contour}
P.~Arbelaez, M.~Maire, C.~Fowlkes, and J.~Malik, ``Contour detection and
  hierarchical image segmentation,'' \emph{IEEE transactions on pattern
  analysis and machine intelligence}, vol.~33, no.~5, pp. 898--916, 2011.

\bibitem{he2015delving}
K.~He, X.~Zhang, S.~Ren, and J.~Sun, ``Delving deep into rectifiers: Surpassing
  human-level performance on imagenet classification,'' in \emph{Proc.~of the
  IEEE Conference on Computer Vision and Pattern Recognition}, 2015, pp.
  1026--1034.

\end{thebibliography}

\end{document}